\documentclass[letterpaper, 10 pt, conference]{ieeeconf}  

\IEEEoverridecommandlockouts                              

\usepackage{mathptmx} 
\usepackage{times} 
\usepackage{amsmath} 
\usepackage{amssymb}  
\usepackage{xcolor}
\usepackage{cite} %
\usepackage{graphicx}

\title{\LARGE \bf
Mars Traversability Prediction: A Multi-modal Self-supervised Approach for Costmap Generation 
}

\author{
    Zongwu Xie\textsuperscript{1}
    Kaijie Yun\textsuperscript{1} 
    Yang Liu\textsuperscript{1}\textsuperscript{*}
    Yiming Ji\textsuperscript{1} 
    Han Li\textsuperscript{1}
\thanks{$^{1}$State Key Laboratory of Robotics and Systems,
        Harbin Institute of Technology, 92 Xidazhijie, Harbin, China
        {\tt\small \{120L030418, jiyiming, 2024112054\}@stu.hit.edu.cn} * Corresponding author: Yang Liu. {\tt\footnotesize liuyanghit@hit.edu.cn}}%
\thanks{This work was supported by by the National Natural Science Foundation of China (Grant No. 5247051643) and the Young Scientists Fund of the National Natural Science Foundation of China (Grant No. 52105014).}
}

\begin{document}

\maketitle
\thispagestyle{empty}
\pagestyle{empty}

\begin{abstract}
We present a robust multi-modal framework for predicting traversability costmaps for planetary rovers. Our model fuses camera and LiDAR data to produce a bird’s-eye-view (BEV) terrain costmap, trained self-supervised using IMU-derived labels. Key updates include a DINOv3-based image encoder, FiLM-based sensor fusion, and an optimization loss combining Huber and smoothness terms. Experimental ablations (removing image color, occluding inputs, adding noise) show only minor changes in MAE/MSE (e.g. MAE increases from ~0.0775 to 0.0915 when LiDAR is sparsified), indicating that geometry dominates the learned cost and the model is highly robust. We attribute the small performance differences to the IMU labeling primarily reflecting terrain geometry rather than semantics and to limited data diversity. Unlike prior work claiming large gains, we emphasize our contributions: (1) a high-fidelity, reproducible simulation environment; (2) a self-supervised IMU-based labeling pipeline; and (3) a strong multi-modal BEV costmap prediction model. We discuss limitations and future work such as domain generalization and dataset expansion.

\end{abstract}

\section{INTRODUCTION}

Autonomous navigation on extraterrestrial terrains, such as the surface of Mars, presents fundamental challenges due to the lack of GPS, sparse terrain priors, and the complex physical properties of the Martian surface. To ensure safe and efficient traversal, planetary rovers must reason about terrain traversability from onboard sensors and generate costmaps that reflect terrain difficulty. However, obtaining reliable traversability labels on Mars is infeasible due to the absence of ground truth and the prohibitive cost of field data collection. As a result, learning-based approaches for costmap generation must rely on realistic simulation and alternative forms of supervision.

Most existing approaches for planetary rover navigation rely either on geometric heuristics derived from Digital Elevation Model (DEM)\cite{gennery1999traversability, 1035370} or handcrafted rules to assign costs based on visual features\cite{manduchi2005obstacle,770402}. While these methods offer basic functionality, they often fail to capture subtle terrain semantics, such as distinguishing between loose dust and compact sand, or differentiating low-lying rocks from traversable gravel. Furthermore, these methods typically ignore dynamic interactions between the robot and the terrain, and thus cannot capture properties like compliance, deformability, or impact roughness that directly affect traversability.

\begin{figure}
    \centering
    \includegraphics[width=\linewidth]{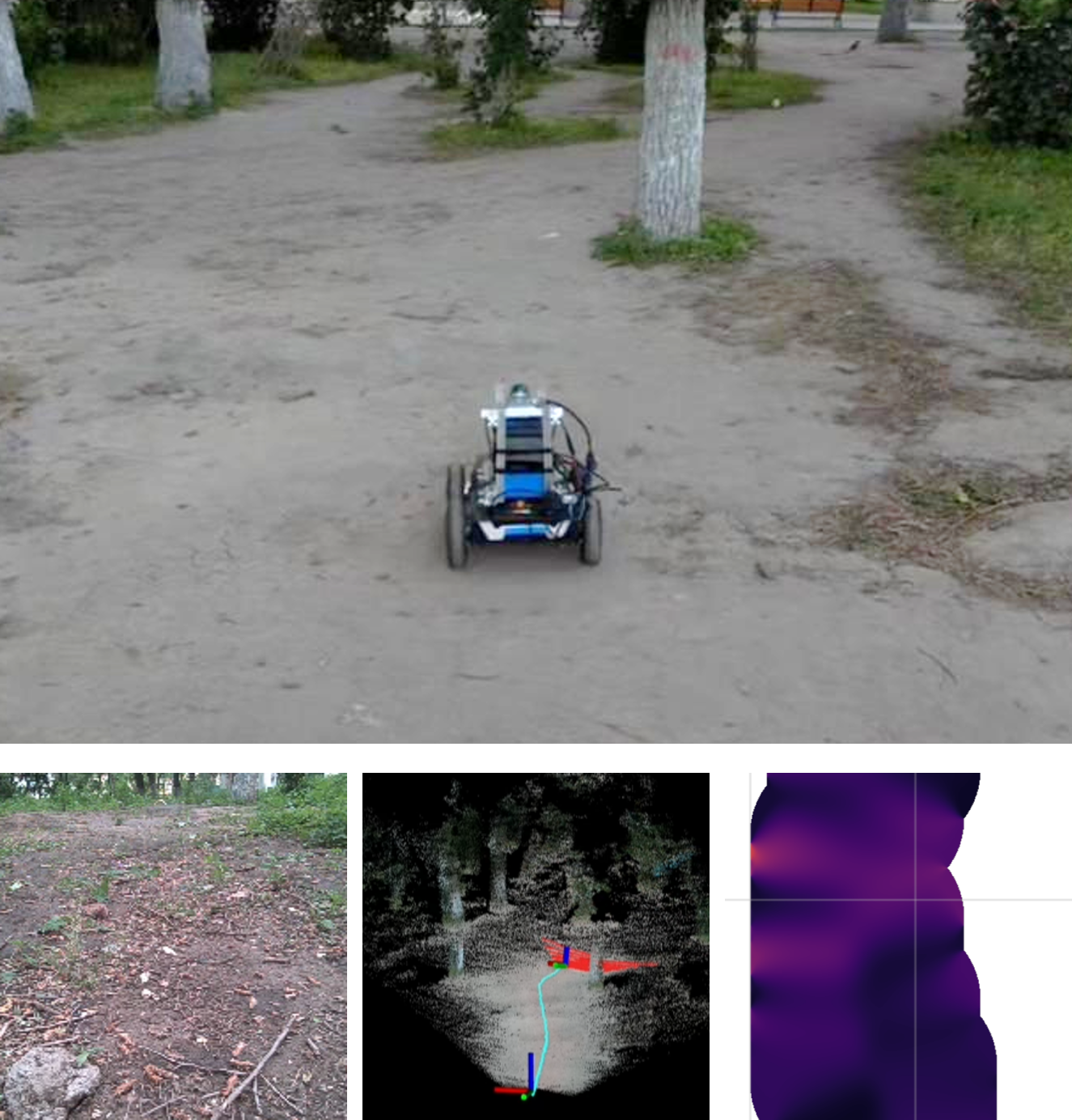}
    \caption{
        \textbf{Real-world data collection and traversability prediction.} The proposed system is validated on a physical four-wheel rover platform operating in outdoor unstructured environments. Top: the rover navigating forest terrain. Bottom: multi-modal perception results, including RGB imagery, LiDAR point cloud with estimated trajectory, and the predicted traversability costmap. This setup demonstrates the feasibility of generating continuous-valued traversability estimates from real-world sensor inputs.
        }
    \label{fig:overall}
\end{figure}

Recent works have explored self-supervised costmap learning using proprioceptive signals such as IMU acceleration\cite{guaman2023hdif}, enabling a more accurate reflection of how the terrain feels to the robot. Others have investigated multi-modal inputs—combining vision, geometry, and semantics—to improve perception robustness across variable terrains\cite{triest2022tartandrive, fan2021learning, zhu2024learning, zhang2024reliable}. Yet, little work has integrated high-fidelity Mars surface models into the costmap learning pipeline, which remains critical for future mission deployment and validation.

In this work, we present a multi-modal traversability learning system that leverages realistic Martian terrain reconstruction and more reliable supervised learning to predict dense, continuous costmaps. Our key contributions are as follows:

\begin{itemize}
    \item \textbf{High-fidelity simulation and dataset.} We build a realistic Mars-terrain simulator and collect paired RGB, LiDAR, and IMU data. This platform ensures reproducibility and allows controlled experiments.
    
    \item \textbf{Self-supervised IMU labeling pipeline.} We introduce a pipeline that converts IMU measurements into continuous BEV cost labels. The IMU-derived cost captures terrain roughness and slippage without human effort, enabling scalable data generation.
    
    \item \textbf{Multi-modal costmap network.} We design a neural network that fuses DINOv3-encoded image features with LiDAR-based BEV features. Our model is robust and produces reliable traversability maps.

\end{itemize}

\section{RELATED WORK}

The field of autonomous robotic navigation in unstructured terrains has evolved significantly over the past decade\cite{10876033,https://doi.org/10.1002/rob.1046,ALKAWI20251756,maturana2017real,240104-0001,app13179877,9636644}. Advances in artificial intelligence have accelerated this trend, making autonomous path planning a focal point. Path planning for Martian rovers, a critical aspect of planetary exploration, often relies on lightweight traversability estimators\cite{9981494,10611227,10611198,10684236} that interpret geometric and visual environmental features. Concurrently, other approaches leverage Digital Elevation Models (DEM)\cite{isprs-archives-XLVIII-3-2024-615-2024} to generate global elevation data and prior maps for evaluating traversability costs.

Recent progress in computer vision has shifted the research focus toward semantic segmentation for traversability estimation in complex off-road environments\cite{long2015fully,noh2015learning}. However, the mapping from human-defined semantic categories to traversability costs remains ambiguous and lacks a direct correspondence. This challenge is exacerbated by the scarcity of Martian visual data, often necessitating the use of heuristic mappings. For instance, while self-supervised methods like the one proposed by Frey et al. \cite{10684236} show promise, their effectiveness can be hindered by reliance on specific sensor suites and an inherently narrow field-of-view, limiting applicability in dynamic Martian terrains.

To overcome the limitations of single-modality approaches, contemporary research has increasingly focused on multimodal learning frameworks that integrate visual\cite{frey2023fasttraversabilityestimationwild,9839522}, geometric\cite{9894664}, and other sensory inputs\cite{9729506}. Some frameworks, such as BADGR\cite{9345970}, employ heuristic label generation for training. However, heuristic-based assessments and those relying solely on geometric data, like the LiDAR-based cost maps from Fan et al.\cite{fan2021learning}, may overlook critical semantic cues essential for robust navigation. Reinforcement learning (RL) offers another powerful paradigm, as demonstrated by Manderson et al.\cite{9196879}, though models trained purely on simulation data face significant challenges in adapting to the physical Martian environment. Other advanced methods derive risk-aware costs from experiential data, such as the Conditional Value-at-Risk (CVaR) approach by Xiaoyi Cai et al.\cite{9982200}. Yet, such methods are often computationally demanding, posing significant challenges for deployment on the resource-constrained hardware of a Mars rover. A persistent challenge across these learning-based methods is the reliance on extensive labeled datasets, a problem that large-scale, self-supervised foundation models like DINOv3\cite{siméoni2025dinov3} aim to mitigate.

A notable trend is the adoption of self-supervised learning, often using IMU signals as a supervisory source, for continuous cost map generation. These methods primarily leverage visual modalities such as RGB imagery\cite{10146445,frey2023fasttraversabilityestimationwild,Shaban2021SemanticTC,9810192,9035009} and point clouds\cite{9738557,rs16091620}. Works on self-supervised costmap learning\cite{guaman2023hdif} and data-driven traversability assessment\cite{zhu2024learning} exhibit conceptual parallels with approaches that learn from proprioceptive feedback. While these methods also focus on quantifying terrain suitability, they often differ in the degree to which they explicitly incorporate dynamic vehicle states (e.g., velocity and acceleration) into their predictive models.

In contrast to the aforementioned works, which often rely on single modalities, heuristic mappings, or computationally intensive models, our primary contribution is a novel and lightweight framework for traversability cost prediction. Our approach uniquely integrates multi-modal data—specifically RGB, point cloud, and semantics—with direct IMU feedback. This is achieved by employing FAST-LIO2\cite{zheng2024fastlivo2fastdirectlidarinertialvisual} and the DINOv3 foundation model to construct the predictive framework. The efficacy and robustness of our method have been rigorously validated through both photorealistic simulations\cite{9636644} and large-scale physical testing in unstructured Martian terrain analogs at elevated velocities.

\section{METHOD}

\subsection{Problem Setup}

Given multi-modal sensory inputs, our objective is to predict a dense 2D traversability costmap in the bird’s-eye view (BEV) coordinate frame of the ego-vehicle. The inputs consist of LiDAR point cloud data and RGB image from single camera. The output is a normalized costmap, where each cell represents the predicted terrain traversal difficulty.

Formally, let the LiDAR point cloud be denoted as $\mathbf{P} = \{ \mathbf{p}_1, \mathbf{p}_2, ..., \mathbf{p}_n \}$, where each point $\mathbf{p}_i \in \mathbb{R}^3$ corresponds to a 3D location $(x_i, y_i, z_i)$ expressed in the LiDAR sensor coordinate frame. The RGB image is denoted as $\mathbf{I} \in \mathbb{R}^{3 \times p_x \times p_y}$, where $p_x$ and $p_y$ represent the image width and height in pixels, respectively.

The target output is a traversability costmap $\mathbf{T} \in [0, 1]^{H \times W}$ in the vehicle’s base (chassis-aligned) BEV coordinate system, where $H$ and $W$ denote the spatial resolution of the BEV grid. Each element $T_{i,j}$ represents the estimated traversability cost at cell $(i,j)$, with $T_{i,j} = 0$ indicating maximum traversability (least cost), and $T_{i,j} = 1$ indicating minimal traversability (highest cost or obstacle).

We aim to learn a function $f$ such that:
$$
f(\mathbf{P}, \mathbf{I}) = \mathbf{T}
$$
where $f$ is implemented as a deep neural network trained to minimize a regression loss between predicted and ground-truth costmaps.

To decompose the problem, we first define a mapping $g$ that projects and encodes the multi-modal input into a shared BEV representation:
$$
    g: (\mathbf{P}, \mathbf{I}) \mapsto \mathbf{B}, \quad \mathbf{B} \in \mathbb{R}^{H \times W \times C}
$$

where $\mathbf{B}$ is a BEV feature map and $C$ is the number of feature channels. The BEV representation aligns all modalities spatially with the output costmap.

Next, we define a cost prediction network $n$ that performs dense regression from the BEV feature map to the final costmap:
$$
n: \mathbf{B} \mapsto \mathbf{T}
$$
Therefore, the complete model can be expressed as a composition of the feature encoder and prediction network:
$$
f = n \circ g
$$

\subsection{Mars Terrain Simulation}

To generate realistic and physically grounded training data for traversability prediction, we construct a high-fidelity simulation environment of the Martian surface using publicly available orbital imagery. Specifically, we leverage digital terrain models (DTMs) and texture maps provided by the High Resolution Imaging Science Experiment (HiRISE) instrument on the Mars Reconnaissance Orbiter. Each DTM provides a georeferenced elevation map of the Martian surface at one meter resolution, while the corresponding texture map offers high-resolution visual appearance data.

To incorporate this data into simulation, we preprocess the HiRISE DTM and texture into formats compatible with the Gazebo physics engine. 
All terrain elements are modeled as rigid bodies in the simulation. While this simplifies the physical representation of surface interaction, it provides sufficient fidelity for simulating wheel-terrain contact under nominal operational conditions. 

A realistic Mars rover model is then placed into the simulated scene. The robot is equipped with a full suite of sensors, including a 3D LiDAR, RGB camera, and inertial measurement unit (IMU). Each sensor is mounted at its calibrated pose relative to the rover base frame, using extrinsic parameters measured from the real system. This ensures that the multi-modal sensor data captured in simulation mirrors the geometric relationships present in a real-world deployment.
\begin{figure}
    \centering
    \includegraphics[width=\linewidth]{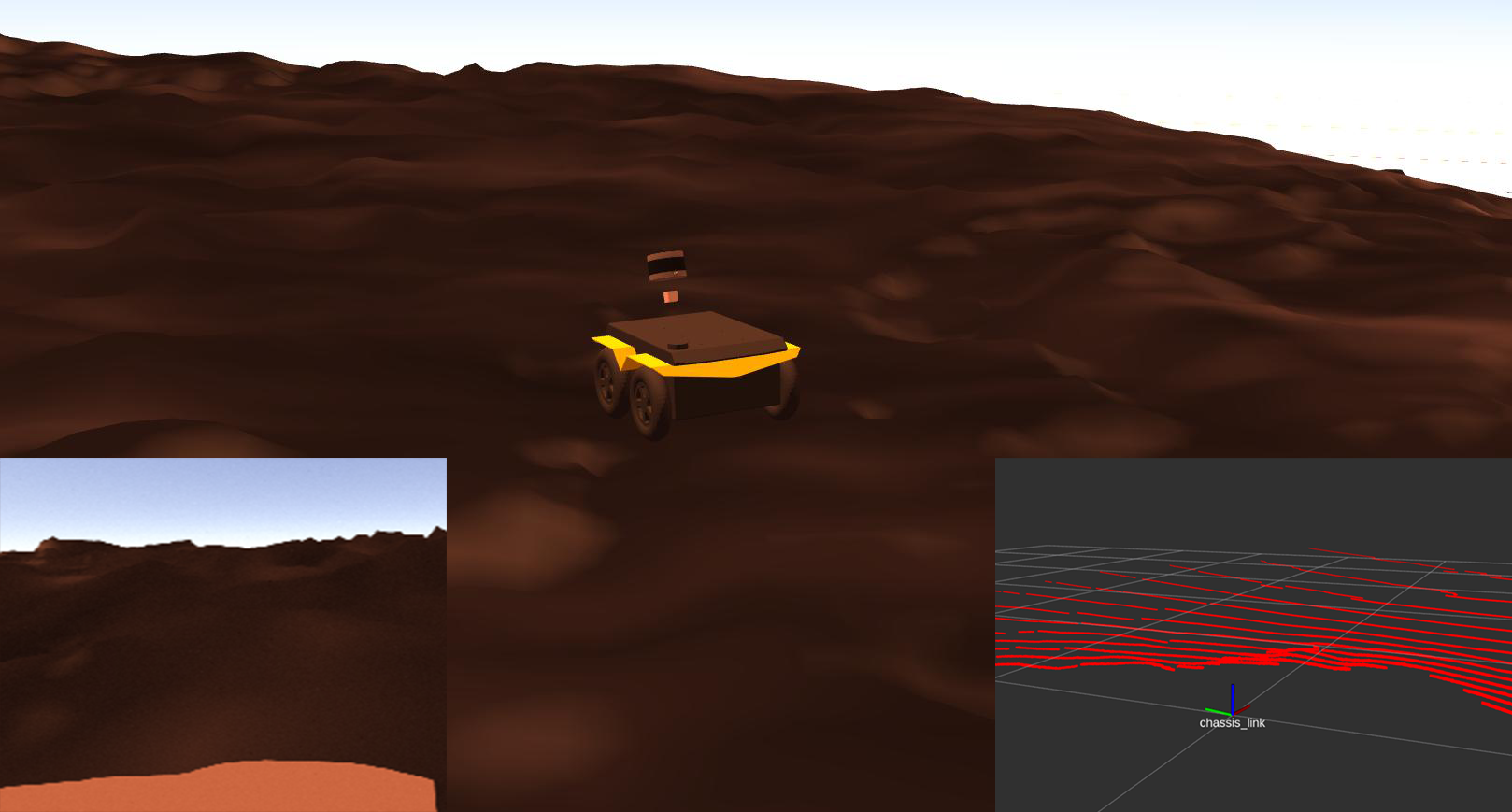}
    \caption{
        \textbf{High-fidelity Mars simulation environment.} A photorealistic simulation environment is constructed using HiRISE digital terrain models and texture maps. The simulated rover is equipped with virtual sensors (LiDAR, camera, IMU), allowing scalable data collection and controllable evaluation across diverse Martian-like terrains. Insets show the elevation profile and simulated LiDAR scan, providing a realistic approximation of Martian surface conditions.
        }
    \label{fig:simulation}
\end{figure}
\subsection{Data Production\label{data production}}

To build a sufficiently large and diverse dataset for training and evaluation, we conduct traversability data collection in two Mars-like terrains. For each terrain map, we design multiple trajectories that not only ensure rover safety—by avoiding hazardous areas such as craters and cliffs—but also maximize coverage of different terrain types. The trajectories are selected to account for both geometric properties (e.g., slope, roughness) and semantic attributes (e.g., rock, sand, soil). In total, we obtain 1,432 trajectory segments that capture heterogeneous Martian surface conditions.

During each data collection run, the rover is initialized at the start of a predefined trajectory and drives forward at a constant speed without human intervention. As it traverses the environment, multi-modal sensor data from the LiDAR, RGB camera, and IMU are continuously recorded in ROS bag files for offline processing. These sensor streams serve as inputs for traversability learning, while the IMU signals are further processed to automatically derive traversability cost (TC) labels.

To assign a TC label to each spatial location, we discretize the trajectory into grid cells of $0.2\,\mathrm{m} \times 0.2\,\mathrm{m}$ resolution, corresponding approximately to the physical footprint of the rover. Within each cell, we extract IMU-derived inertial features and compute a cost score reflecting the local traversability difficulty. At each response, the IMU reports linear acceleration $\mathbf{a}_i = [a_x, a_y, a_z]^\top \in \mathbb{R}^3$ and angular velocity $\boldsymbol{\omega}_i = [\omega_x, \omega_y, \omega_z]^\top \in \mathbb{R}^3$, both expressed in the ego frame of the rover. For a given spatial segment indexed by $i$, we define:
\begin{align}
    \Delta \mathbf{a}_i &= \mathbf{a}_{i+1} - \mathbf{a}_i \nonumber\\ 
    \Delta \mathbf{p}_i &= \mathbf{p}_{i+1} - \mathbf{p}_i, \quad \Delta s_i = \|\Delta \mathbf{p}_i\|_2\nonumber
\end{align}

where $\mathbf{p}_i \in \mathbb{R}^3$ is the rover position at timestamp $i$, and $\Delta s_i$ is the Euclidean distance between consecutive points. To avoid division by zero, we regularize small $\Delta s_i$ by setting a minimum threshold $\epsilon$. The acceleration magnitude and spatial-domain jerk are given by:
\begin{align}
    a_{\text{mag}}(i) &= \|\mathbf{a}_i\|_2\nonumber \\
    \mathbf{j}_i &= \frac{\Delta \mathbf{a}_i}{\Delta s_i}, \quad j_{\text{mag}}(i) = \|\mathbf{j}_i\|_2\nonumber
\end{align}

The cumulative angular change per unit distance is computed as:
\begin{align}
    \theta_{\text{cum}} = \frac{\sum_{i=1}^{T-1} \|\boldsymbol{\omega}_i\|_2 \cdot \Delta s_i}{\sum_{i=1}^{T-1} \Delta s_i}\nonumber
\end{align}

We define the final traversability cost $TC$ for each grid cell as:
\begin{align}
    TC = & \; w_1 \cdot \sqrt{\frac{1}{T} \sum_{i=1}^{T} a_{\text{mag}}(i)^2} \nonumber \\
         & + w_2 \cdot \theta_{\text{cum}} 
           + w_3 \cdot \sqrt{\frac{1}{T-1} \sum_{i=1}^{T-1} j_{\text{mag}}(i)^2} \nonumber
\end{align}

where $T$ is the number of samples in the current spatial cell, and $w_1 = w_2 = w_3 = 1$ are scalar weights set in our experiments.

\begin{figure*}
    \begin{center}
        \includegraphics[width=\linewidth]{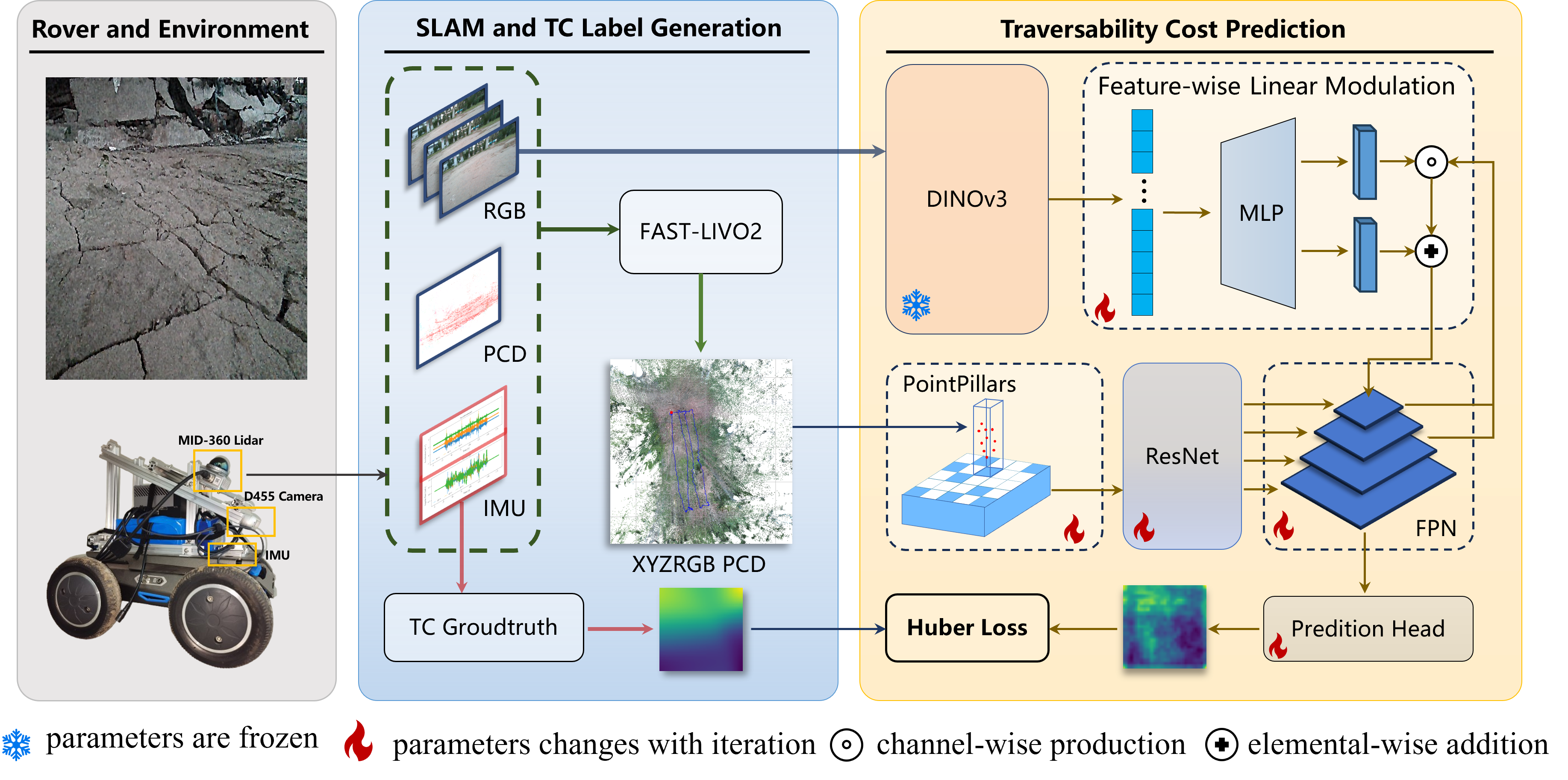}
    \end{center}
    \caption{
        \textbf{Overview of the proposed traversability learning framework.} The pipeline consists of three main modules: (a) rover and environment, where a four-wheel rover equipped with LiDAR, RGB camera, and IMU is deployed in Mars-analog terrains; (b) SLAM and TC label generation, where multi-modal sensor data are temporally aligned via FAST-LIO2 and IMU-based metrics are used to derive traversability cost ground truth; and (c) traversability cost prediction, where RGB features from DINOv3 are fused with point cloud features from PointPillars via feature-wise linear modulation, and the fused BEV features are decoded by ResNet–FPN into a dense costmap, optimized with Huber and smoothness losses.
        }
    \label{fig:model_arch}
\end{figure*}

Unlike traditional IMU-based methods that rely solely on vertical acceleration to infer bumpiness or pitch, our approach incorporates the full 3D inertial response---including lateral slip and roll---to capture a more comprehensive estimate of terrain difficulty. Furthermore, all inertial computations are performed in the spatial domain rather than in the time domain, which reduces sensitivity to variable rover speed and ensures that cost estimation remains consistent across different trajectories.

To ensure spatial continuity of the discrete TC labels, we adopt a Bayesian kernel interpolation method~\cite{shan2018bayesian}, which upsamples the original $0.2\,\mathrm{m}$ grid costs to a finer $0.01\,\mathrm{m}$ resolution. 
Let the known points be $\left(x_n, y_n, TC_n\right)$ and the target query point be $\left(x,y\right)$. 
The distance between them is denoted as $d_n$. 
We define a sparse kernel function:
\begin{align*}
K(d;r) =
\begin{cases}
  \frac{2 + \cos \frac{2\pi d}{r}}{3}(1-\frac{d}{r}) + \tfrac{\sin\frac{2\pi d}{r} }{2\pi}, & d \leq r, \\
  0, & d > r ,
\end{cases}
\end{align*}
where $r$ is the kernel radius (set to $1.0\,\mathrm{m}$ in our experiments). 
The interpolated TC value is then given by
\begin{equation*}
\widehat{TC}(x,y) = 
\frac{\sum_{n=1}^N K(d_n; r)\,TC_n}{\sum_{n=1}^N K(d_n; r)} .
\end{equation*}


\subsection{Costmap Prediction Model}

In this work, we propose a traversability cost (TC) learning framework based on multi-modal inputs. The central idea is to fuse geometric structure from point clouds with texture and appearance cues from RGB images to directly generate a costmap that reflects terrain traversability. Unlike conventional methods that rely on semantic segmentation followed by remapping into costs, our approach learns an end-to-end mapping from raw multi-modal data to the costmap. This design simplifies the pipeline and improves generalization ability across diverse terrains. The overall architecture of the prediction model is illustrated in Fig.~\ref{fig:model_arch}.

\paragraph{Input Representation}
The inputs consist of point cloud and RGB image data. The point cloud is denoted as 
\begin{equation*}
    P = \{(x_i, y_i, z_i, r_i, g_i, b_i)\}_{i=1}^N,
\end{equation*}
where $N$ is the number of points. The image input is represented as 
\begin{equation*}
    I \in \mathbb{R}^{B \times H \times W \times 3},
\end{equation*}
with $B$ denoting the batch size (equivalent to the number of input patches), and $H, W$ the image dimensions. Temporal and spatial alignment between point clouds and images is ensured using the FAST-LIVO2 framework~\cite{zheng2024fastlivo2fastdirectlidarinertialvisual}.

\paragraph{Image Feature Extraction}
For image encoding, we employ the self-supervised DINOv3 model~\cite{siméoni2025dinov3}, which has been pre-trained on large-scale datasets and provides strong visual representation and cross-domain generalization. An input image $I$ is mapped into a high-level feature embedding:
\begin{equation*}
    F_I = \mathrm{DINOv3}(I), \quad F_I \in \mathbb{R}^{B \times d},
\end{equation*}
where $d=384$ is the embedding dimension. Instead of decoding $F_I$ into semantic maps, we directly combine these high-level embeddings with point cloud features, thereby implicitly leveraging semantic and appearance information from the image.

\paragraph{Point Cloud Encoding}
Point cloud encoding is performed using the PointPillars method~\cite{lang2019pointpillars}, which converts unordered 3D points into structured pillar representations in bird’s-eye view (BEV) space. The point cloud $P$ is partitioned into vertical pillars, and features within each pillar are aggregated via a multi-layer perceptron (MLP). A ReLU activation selects the strongest response in each pillar, producing a compact feature representation. This results in a pseudo-image:
\begin{equation*}
    V = \mathrm{PillarPillar}(P), \quad V \in \mathbb{R}^{H' \times W' \times C},
\end{equation*}
where $H', W'$ denote the BEV grid size, and $C$ the feature channels.  
This BEV pseudo-image is then fed into a ResNet-50 backbone~\cite{he2016deep} followed by a Feature Pyramid Network (FPN)~\cite{lin2017feature}, yielding multi-scale features:
\begin{equation*}
    F_{\text{bev}} = \{p_1, p_2, p_3, p_4\}.
\end{equation*}

\paragraph{Image-Point Cloud Feature Fusion}
Since DINOv3 features are linear embeddings, we adopt Feature-wise Linear Modulation (FiLM)~\cite{perez2018film} for feature fusion. The image feature $F_I$ is first projected by an MLP into modulation parameters $(\gamma, \beta)$, which then modulate the BEV features channel-wise:
\begin{equation*}
    \widetilde{F}_{\text{bev}} = \gamma(F_I) \odot F_{\text{bev}} + \beta(F_I),
\end{equation*}
where $\odot$ denotes channel-wise multiplication. In practice, we apply modulation only to higher-level features $p_3, p_4$, which yields strong performance while maintaining efficiency.

\paragraph{Prediction Head}
The fused feature maps are passed through a lightweight prediction head consisting of two convolutional layers. The first layer reduces the FPN features to 64 channels followed by ReLU activation, while the second layer outputs a single-channel costmap. A Sigmoid activation normalizes the predictions into $[0,1]$, and bilinear upsampling restores the output to the original resolution:
\begin{equation*}
    \widehat{C} = f_\theta(\widetilde{F}_{\text{bev}}), \quad \widehat{C} \in \mathbb{R}^{H' \times W'}.
\end{equation*}

\paragraph{Supervision and Loss Functions}
The ground-truth traversability costs are derived from IMU signals as described in Sec.~\ref{data production}. They provide continuous and objective supervision of terrain roughness. We adopt a combination of Huber loss and a smoothness regularizer.

The Huber loss is defined as:
\begin{equation*}
\mathcal{L}_{\text{Huber}} =
\begin{cases}
  0.5(\widehat{C}_{i,j} - C_{i,j})^2, & |\widehat{C}_{i,j} - C_{i,j}| < \delta, \\
  \delta\left(|\widehat{C}_{i,j} - C_{i,j}| - 0.5\delta\right), & \text{otherwise},
\end{cases}
\end{equation*}
where $\delta$ is the transition threshold.

The smoothness loss is defined as:
\begin{align*}
\mathcal{L}_{\mathrm{Smooth}} &= \lambda_s \Bigg(
\frac{1}{(H-1)W}\sum_{i=1}^{H-1}\sum_{j=1}^{W}\big|\widehat{C}_{i+1,j}-\widehat{C}_{i,j}\big| \notag\\
&\qquad
+ \frac{1}{H(W-1)}\sum_{i=1}^{H}\sum_{j=1}^{W-1}\big|\widehat{C}_{i,j+1}-\widehat{C}_{i,j}\big|
\Bigg)
\end{align*}
where $\lambda_s=0.1$ in our experiments.

The final objective is:
\begin{equation*}
    \mathcal{L} = \mathcal{L}_{\text{Huber}} + \mathcal{L}_{\text{Smooth}}
\end{equation*}

\paragraph{Training and Optimization}
The model is trained end-to-end with Adam optimizer, with an initial learning rate of $1\times10^{-4}$ and a batch size of 8. Data augmentation, including random rotation, translation, and Gaussian noise, is applied to both images and point clouds to improve robustness to noisy and diverse Martian terrain conditions.

\section{EXPERIMENT}
\subsection{Dataset}
The public dataset were utilized in this study: HiRISE dataset.  


The \textbf{HiRISE dataset} was used to construct high-fidelity Martian terrain simulations. High Resolution Imaging Science Experiment (HiRISE) digital terrain models (DTMs) are generated from stereo image pairs acquired at different viewing angles, with resolutions of 0.25–0.5~m/pixel, corresponding to terrain post spacing of approximately 1–2~m and vertical accuracy on the order of tens of centimeters. 

\subsection{Hardware Devices}
The experimental platform consists of a four-wheel differential-drive rover, a multi-sensor perception suite, and an onboard computing unit.

\textbf{LiDAR:} A Livox MID-360 is used, operating at a wavelength of 905~nm. It provides a horizontal field of view (FoV) of $360^\circ$ and a vertical FoV of $-7^\circ$ to $52^\circ$, with a maximum range of 70~m at 80\% reflectivity. The ranging accuracy is $\leq$2cm at 10m, delivering an average point cloud density of 200,000 points/s at a frame rate of 30~Hz.

\textbf{Camera:} An Intel RealSense D455 RGB camera is employed, operating at a resolution of $1280\times720$ and an FoV of $90^\circ \times 65^\circ$. The image stream is published at 30~Hz.

\textbf{IMU:} A WHEELTEC N100 inertial measurement unit is used, providing acceleration and angular velocity measurements at 200~Hz.

\textbf{Mobile platform:} A custom-built four-wheel differential-drive rover from HEXMAN serves as the base platform for sensor mounting. Both the LiDAR and camera are mounted with an approximately $15^\circ$ downward tilt to maximize ground coverage.

\textbf{Computing unit:} An NVIDIA Jetson AGX Orin is mounted on the rover for real-time sensor data acquisition, preprocessing, and neural network inference. 

All model training and simulation were conducted on a workstation equipped with four NVIDIA GeForce RTX~4090 GPUs, while real-world experiments were executed entirely onboard the Jetson AGX Orin.

\subsection{Ablation Study}

The most direct way to validate the effectiveness of the TC prediction model is to use the predicted risk maps for path planning, and then compute the actual accumulated TC values of the planned trajectories for comparison. However, random variations introduced by the path planning algorithm make it challenging to isolate the influence of the prediction model.  

To eliminate this source of randomness, we define fixed test routes in real-world environments. For each route, the ground-truth accumulated TC value is computed from IMU measurements. The robot traverses each route at different constant velocities while collecting multi-modal sensory data (RGB images and LiDAR point clouds). The collected data is then fed into different model variants, and the Mean Absolute Error (MAE) and Mean Squared Error (MSE) between the predicted and ground-truth TC values are computed.

To evaluate the contribution of different input modalities and to assess the robustness of the proposed TC prediction model, we design 5 groups of ablation experiments. It should be noticed that all test ablations use full model weight. Results ablation groups are summarized in Table~\ref{tab:ablation_res}.

\begin{itemize}
    \item \textbf{No-Color-PointCloud:} Remove the RGB color channels of the point cloud (i.e., use only geometry). Image encoder and FiLM remain.
    \item \textbf{No-Image-Encoder:} Remove the DINOv3 encoder and FiLM. Instead, we only map raw image colors onto points without learned modulation.
    \item \textbf{Occluded-Image:} At test time, randomly mask portions of the input image to simulate camera occlusion.
    \item \textbf{Sparse-PointCloud: } Randomly drop 30\% of LiDAR points to simulate low-resolution sensing.
    \item \textbf{Noisy-Inputs: } Add Gaussian noise to both image pixels and point cloud positions to test sensor noise robustness.
    
\end{itemize}

\begin{table}[h]
\centering
\caption{Ablation results}
\label{tab:ablation_res}
\begin{tabular}{lccccl}
\hline
\textbf{Experiment} & \textbf{MAE} &  \textbf{MSE} \\
\hline
\textbf{Full model (baseline)} & 0.0775 & 0.0118 \\
No-Color-PointCloud (train) & \textbf{0.0637} & \textbf{0.0084}  \\
No-Image-Encoder (train) & 0.0769 & 0.0115 \\
Occlude Image (test) & 0.0775 & 0.0118  \\
Sparse Point Cloud (test) & 0.0915 & 0.0159  \\
Add Gaussian Noise (test) & 0.0860 & 0.0143  \\
\hline
\end{tabular}
\end{table}

These results indicate the model relies primarily on LiDAR geometry. Removing image color or even the entire image feature branch causes negligible or slight improvement, suggesting that the IMU-derived labels contain little semantic information to learn from. In particular, the cost labels generated from IMU largely reflect physical bumps/slippage, which in our dataset (relatively uniform rocky terrain) did not strongly correlate with visual appearance. As a result, the network learned to prioritize geometric cues, and the image branch acted mostly as a minor conditioning signal.

\subsection{Analysis}
The small performance differences point to two main factors. First, the IMU cost labels are insensitive to semantics: subtle effects like wheels slipping on sand vs gravel produce minimal IMU changes in our data, so the “ground truth” cost is dominated by overall terrain ruggedness. This means image features (semantic cues) had little additional signal, and in fact sometimes slightly hurt generalization. Second, the dataset is relatively small and homogeneous, so the complex model can overfit. We observed that the full model converged faster (training error lower) than the simpler models, indicating overfitting, but all achieved similar validation error. Adding data diversity in the future would allow the image branch to show more benefit.

Importantly, despite these ablations, all MAE values remain low (around 0.06–0.09 on a [0,1] scale), demonstrating strong robustness. The baseline and image-occluded models share identical error, and even under noisy/sparse inputs the error increases only modestly. This robustness is encouraging for real-world deployment, where sensor faults may occur.

These findings are consistent with the idea that our architecture can effectively fuse modalities, but also that training labels and data distribution ultimately dictate what the model learns.

\section{CONCLUSION}
We have presented a multi-modal learning framework for Mars-rover traversability, emphasizing a reproducible pipeline and robustness. By combining a DINOv3 image encoder with a LiDAR BEV backbone and FiLM fusion, we achieve reliable costmap predictions. Our self-supervised IMU-based labeling and high-fidelity simulation environment facilitate training without manual annotations.

Unlike prior work that emphasized large performance gains, our experiments show only incremental improvements when adding image information. This highlights that the key contributions are: (1) the public, high-fidelity simulation environment and dataset, which can accelerate future research; (2) the self-supervised IMU-cost labeling pipeline enabling scalable data generation; and (3) a robust multi-modal BEV costmap network that generalizes well under input corruptions.

We also openly discuss limitations. The learned model is currently constrained by the IMU-label characteristics and a relatively narrow training set. Future work will focus on domain generalization (e.g. transferring from simulated Mars scenes to real rover data) and expanding the dataset with more terrain types and conditions. For instance, incorporating different lighting or atmospheric conditions, or adding semantic terrain categories (rocks, sand, cliffs) could improve the use of visual features. We also plan to explore unsupervised domain adaptation and more advanced architectures to better leverage multi-sensor cues. Overall, we provide a solid baseline and tools for planetary traversability learning, with an emphasis on reproducibility and practical robustness.

\bibliographystyle{IEEEtran}
\bibliography{references}

\addtolength{\textheight}{-12cm}   







\end{document}